\documentclass[final]{cvpr}

\usepackage{times}
\usepackage{epsfig}
\usepackage{graphicx}
\usepackage{amsmath}
\usepackage{amssymb}
\usepackage{subcaption}

\usepackage[pagebackref=true,breaklinks=true,colorlinks,bookmarks=false]{hyperref}

\begin{document}

%%%%%%%%% TITLE
\title{End-to-End Human Object Interaction Detection with HOI Transformer}

\author{Cheng Zou\thanks{Equal contribution.},
% \and 
Bohan Wang$^*$, 
% \and 
Yue Hu, 
% \and 
Junqi Liu,
% \and
Qian Wu,
% \and
Yu Zhao,
% \and
Boxun Li, \\
% \and
Chenguang Zhang,
% \and
Chi Zhang,
% \and
Yichen Wei,
% \and
Jian Sun\\

MEGVII Technology\\
{\tt\small \{zoucheng,wangbohan,huyue,liujunqi,wuqian,zhaoyu03,liboxun}\\ 
{\tt\small zhangchenguang,zhangchi,weiyichen,sunjian\}@megvii.com }
}
% \and

\maketitle
%%%%%%%%% ABSTRACT

\begin{abstract}
We propose HOI Transformer to tackle human object interaction (HOI) detection in an end-to-end manner. Current approaches either decouple HOI task into separated stages of object detection and interaction classification or introduce surrogate interaction problem. In contrast, our method, named HOI Transformer, streamlines the HOI pipeline by eliminating the need for many hand-designed components. HOI Transformer reasons about the relations of objects and humans from global image context and directly predicts HOI instances in parallel. A quintuple matching loss is introduced to force HOI predictions in a unified way. Our method is conceptually much simpler and demonstrates improved accuracy. Without bells and whistles, HOI Transformer achieves $26.61\% $ $ AP $ on HICO-DET and $52.9\%$ $AP_{role}$ on V-COCO, surpassing previous methods with the advantage of being much simpler. We hope our approach will serve as a simple and effective alternative for HOI tasks. Code is available at \url{https://github.com/bbepoch/HoiTransformer}.
\end{abstract}

%%%%%%%%% BODY TEXT

\section{Introduction}
% 1. HOI 很重要
Human-Object Interaction (HOI) detection plays an important role in the high level human-centric scene understanding, and has attracted considerable research interest recently. 
The HOI research can also contribute to other tasks, such as action analysis, weakly-supervised object detection, and visual question answering, etc.

\begin{figure}[t]
\centering
\includegraphics[scale=0.55]{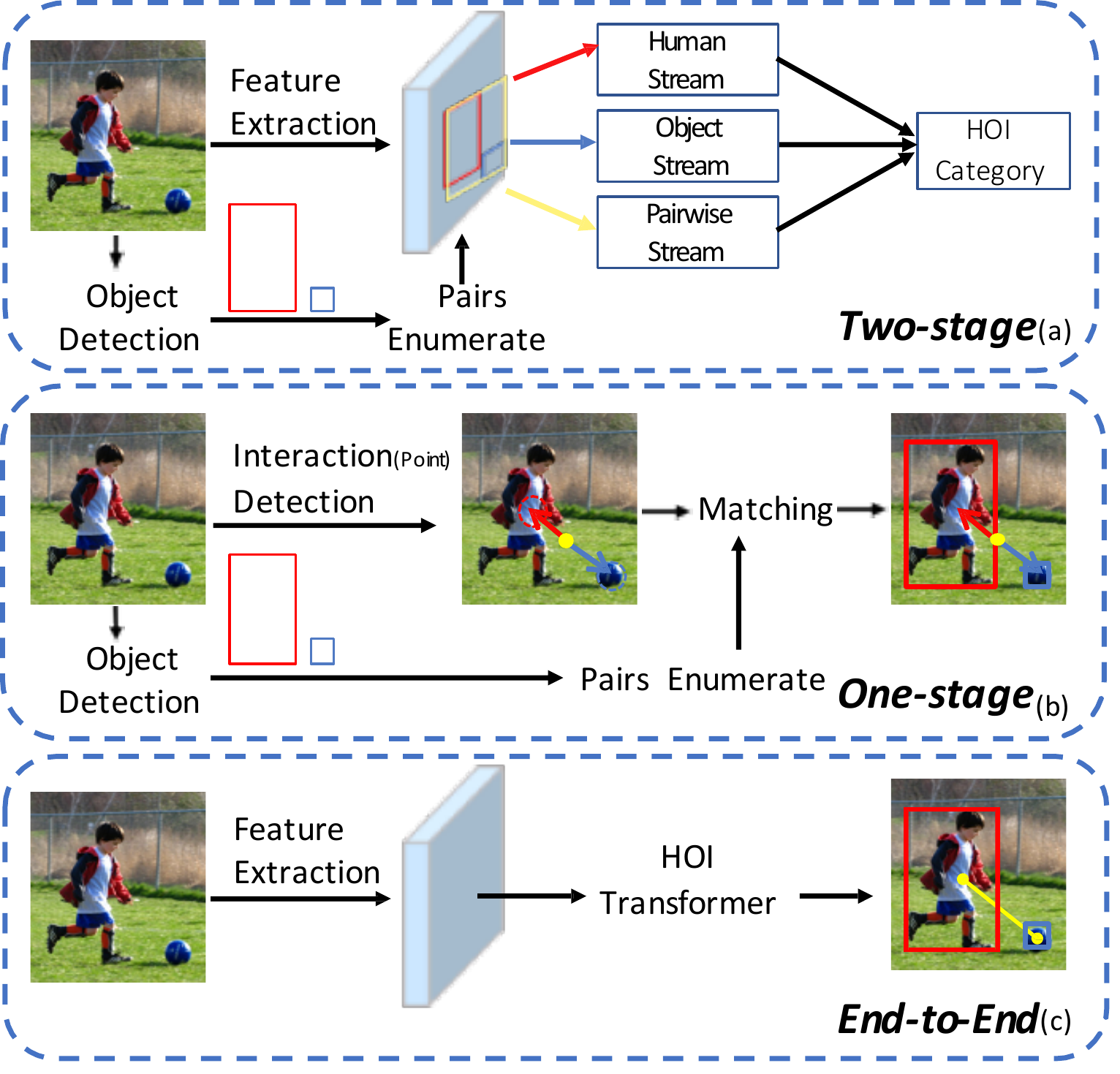}
\caption{Comparison of recent approaches on HOI detection. (a) two-stage methods, typically use pre-trained detectors to generate human, object proposal and enumerate all (human, object) pairs, followed by a multi-stream architecture to clarify interaction. (b) one-stage methods, detect interaction point/box and object proposals simultaneously, followed by complex matching process to assign interactions to object pairs. (c) our end-to-end method, input an image and predict HOI instances directly.}
\label{fig:12shot}
\end{figure}

The goal of HOI detection aims at localizing human and object, as well as recognizing the interaction between them. 
Previous studies~\cite{gao2018ican, eccv2020dualgraph, kaiming2018detecting, eccv2020keycues,liao2020ppdm,eccv2020uniondet} present promising results on HOI detection by decouple this task into object detection and interaction classification (Fig.~\ref{fig:12shot}(a)). More specifically, human and object detection results is first obtained by pre-trained object detector, then interaction classification is conducted on the pair-wisely combined human-object proposals. The limitations of these methods are mainly caused by the \emph{separated} two stages. The independent optimization on two sub-problems may lead to sub-optimal solution. The generated human-object proposals have relative low quality for interaction classification~\cite{liao2020ppdm}, because only object-level confidence has been taken into account. Moreover, all pair-wise proposals need to be processed, which brings large redundant computation cost.

More recent approaches~\cite{wang2020irnet, eccv2020uniondet, liao2020ppdm} have introduced a surrogate interaction detection problem to optimize HOI detection indirectly (Fig.~\ref{fig:12shot}(b)). Firstly, the interaction proposal has been pre-defined based on human priors. For example, UnionDet~\cite{eccv2020uniondet} defines the interaction proposal as union box of the human and object box. PPDM~\cite{liao2020ppdm} uses the center point between human and object as interaction point. Secondly, the human, object and interaction proposals are detected in parallel. Finally, each interaction result is assigned to one (human, object) pair based on pre-defined matching strategy in post processing. 
However, such definition of interaction proposal are not always valid under different circumstance and make the pipeline more complex and costly in computation.

% 说transformer好 好的hoi模型应该解决这2点 建模关系 不需要预测物体 直接出结果

% 提纲

For HOI detection, how to capture the dependencies, especially long range, between human and object in the image space is the main problem. The above methods used complex but sub-optimal strategies, i.e. decouple into two-stages or introduce surrogate proposals to empower models the ability of capturing dependencies. However, the transformer network~\cite{transformer} is designed to exhaustively capture the long range dependencies, which inspire us to address the problem with transformer.

In this paper, we propose a new architecture  to directly predict the HOI instance, i.e. (human, object, interaction), in an \emph{end-to-end} manner.
Our method consists of two parts, a transformer encoder-decoder architecture and a quintuple HOI matching loss. 
The architecture first use CNN backbone to extract high-level image features, then the encoder is leveraged to generate global memory feature, which models the relation between the image feature explicitly. Next the global memory from encoder and the HOI queries are send to decoder to generate the output embeddings. Finally, a multi-layer perception is used to predict HOI instances based on the output embeddings of decoder. Meanwhile, a quintuple HOI matching loss is proposed to supervise the learning of HOI instance prediction. Our method achieves state-of-the-art results on different challenging HOI benchmarks.

% The major contributions of this are as follows:
% \begin{itemize}
% \item[1)] We propose a simple yet effective transformer encoder-decoder to architecture to directly predict HOI instance without complex post-process.
% \item[2)] We propose a quintuple matching loss for HOI to optimize the model in a unified way.   
% \item[3)] Our approach achieves state-of-the-art performance on two challenging HOI detection benchmarks: V-COCO and HICO-DET.
 
% \end{itemize}

\section{Related work}
\subsection{Two-Stage HOI Detection}
Modern two-stage HOI detection methods usually consists of an object detector in the first stage and an interaction classifier in the second stage. More specifically, In the first stage, a fine-tuned object detector is used to get the humans and objects bounding boxes and class labels. In the second stage, a multi-stream architecture is used to predict the interactions for each human-object pair. 

Typically there are three streams in the mentioned multi-stream interaction classifier: human stream, object stream, and pairwise stream. Both human stream and object stream usually encode visual features for human and object boxes respectively~\cite{gao2018ican}. In FCMNet~\cite{eccv2020keycues}, object visual feature is replaced by word embedding for the reason that detailed visual appearance of the object is often not crucial for the interaction category. Besides visual features, Bansal et. al~\cite{bansal2020detecting} introduced word embedding in human stream for feature augmentation. PDNet~\cite{eccv2020polysemy} introduced word embedding for all the streams to get language prior-guided channel attention and feature augmentation. Plenty of researches have been done on the pairwise stream. This stream usually encodes the relationship between the human and object. A two-channel binary image representation is first advocated in iCAN~\cite{gao2018ican} to encode the spatial relation, but in FCMNet~\cite{eccv2020keycues}, a fine-grained version from human parsing is proposed to amplify the key cues. Apart from spatial relation, graph neural networks in DRG~\cite{eccv2020dualgraph}, CHG~\cite{eccv2020hetegraph}, RPNN~\cite{zhou2019relation} were proposed to explicitly model the interactions between human and objects, which sure improved the model’s representation capability. % Except for the above three streams, extra ones such as human pose stream~\cite{li2019tin,wan2019pose}, Action Co-occurrence Priors~\cite{eccv2020actioncoprior},  and context stream~\cite{} were also proved to work well.

Auxiliary models can be easily introduced to two-stage pipeline to help improving HOI, e.g. human pose feature, human body-part~\cite{li2020pastanet}, language model~\cite{peyre2019detecting} and graph model~\cite{xu2019learning}, etc. Interestingly Bansal et.al~\cite{bansal2020detecting} and Hou et.al~\cite{eccv2020visualcomlear} introduced feature level augmentation, which is proved to be effective to HOI. However, these methods suffer from heavy complexity and low efficiency due to the sequential and separated two-stage architecture. 

\begin{figure*}[th]
\centering
\includegraphics[scale=0.5]{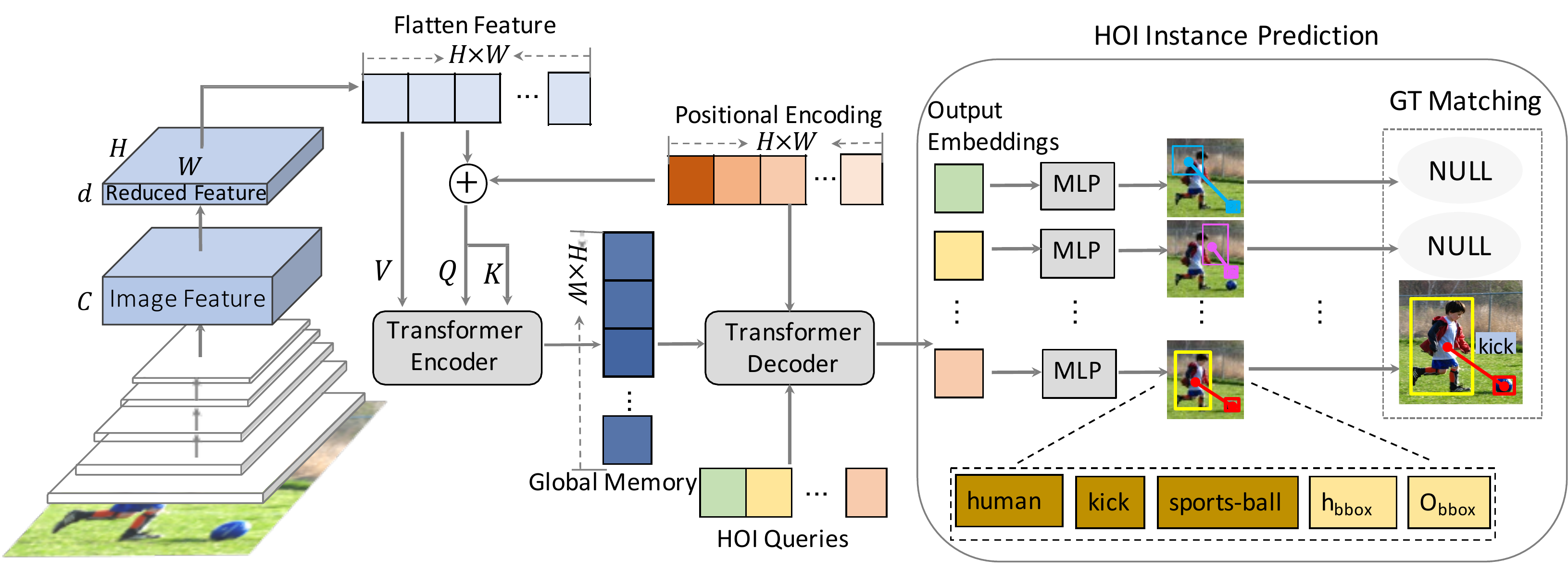}
\caption{Overall architecture. Our model first use a CNN backbone to extract visual feature from input image. Then the feature is reduced in channel-dimension, flatten in spatial-dimension and complemented by positional encoding. The transformer encoder generates the global memory feature based on $K,Q,V$. The transformer decoder transforms $N$ learnt positional embeddings~(denoted as HOI queries) into $N$ output embeddings. Finally, the multi-layer perception~(MLP) predict the quintuple HOI instance based on the embeddings. The HOI instances are directly output simultaneously. % In the training process, our HOI Matching assign each predicted HOI instance to one ground-truth based and is optimized end-to-end. During inference time, the HOI instance is directly output without post-process.
}
\label{fig:frame}
\end{figure*}

\subsection{One-Stage HOI Detection}
InteractNet~\cite{kaiming2018detecting} may be one of the earliest detection-based methods but it needs cascaded inference. PPDM~\cite{liao2020ppdm} and IPNet~\cite{wang2020irnet} treated HOI as a point detection problem and directly detect interactions in a one stage manner by introducing a novel definition of interaction point. Further, PPDM predicts both object detection and HOI detection in a unified CenterNet-based~\cite{zhou2019objects} model. In UnionDet~\cite{eccv2020uniondet}, HOI detection is regarded as a union box detection problem and based on the popular RetinaNet~\cite{lin2017focal}, another unified one stage HOI detection model is proposed, an extra union branch for detecting union box is added parallel to the conventional object detection branch. 

% The one-stage HOI detection models are usually embedded in detection framework, and trained with image data only. So, it is not that flexible, and it is not  easy for them to use augmentations just like VCL~\cite{eccv2020visualcomlear}. 

Compared to two-stage methods, the pipeline becomes simpler, faster, more efficient and easier to deploy for real world applications. 
However, one-stage methods still need complex post processing to group object detection results and the interaction predictions.

\subsection{End-to-End Object Detection}
Russell ~\cite{ng} proposed an end-to-end people detection method by LSTM-based encoder-decoder, which is an auto-regressive model that predicts the output sequence one element at a time. DETR~\cite{detr} improved it by by replacing LSTM with transformer, which decodes N objects in parallel by leveraging the recent transformers with parallel decoding~\cite{oord2018parallel,gu2017non,ghazvininejad2019mask}. Both methods use Hungarian algorithm to match the ground truths and predictions though different matching costs are used.
Unlike traditional object detectors, the end-to-end methods, usually have an NMS free architecture, and to make this reality, a good one-to-one matching strategy for duplicates reduction is important, and Hungarian matching seems to be a better choice so far.

\section{Method}
% 我们方法的如何直接输出triplet-wise的 direct relation prediction: 完善方法；往多的写
\subsection{Overview}
Different from previous methods, we solve human-object interaction detection in an end-to-end manner both in training and inference: input an image and then output the HOI relations directly, without any post processing. The proposed method consists of two main parts, an end-to-end transformer encoder-decoder architecture and a quintuple HOI instance matching loss. % In this part, we will first briefly give an overview of the transformer encoder-decoder architecture, and then detail the MLE matching loss.

\subsection{Network Architecture}
The proposed architecture illustrated in Fig.~\ref{fig:frame} consists of three main parts: (i) a backbone to extract visual feature from the input image, (ii) a transformer encoder-decoder to digest backbone feature and produce output embeddings, and (iii) a multi-layer perception (MLP) to predict HOI instances.

\noindent\textbf{Backbone}: A CNN backbone is used to extract visual feature from the input image. First, a color image is fed into the backbone and generate a feature map of shape $(H, W, C)$ which contains high level semantic concepts. A $1 \times 1 $ convolution layer is used to reduce the channel dimension from $C$ to $d$. A flatten operator is used to collapse the spatial dimension into one dimension. After that, a feature map of shape $[H \times W, d]$ is obtained, denoted as \emph{flatten feature} in Fig.~\ref{fig:frame}. The spatial dimension transformation is important because the following transformer encoder requires a sequence as input, thus the feature map can be interpreted as a sequence of length $H \times W$, and the value at each time step is a vector of size $d$. We use ResNet~\cite{he2016deep} as our backbone and reduce the dimension of  feature conv-5 from $C=2048$ to $d=256$.

\noindent\textbf{Encoder}: The encoder layer is built upon standard transformer architecture with a multi-head self-attention module and a feed-forward network~(FFN). Theoretically the transformer architecture is permutation invariant. To enable it distinguish relative position in the sequence, position encoding~\cite{parmar2018image,bello2019attention} is added to the input of each attention layer. The sum of flatten feature and \emph{positional encoding} is fed into the transformer encoder to summarize global information. The output of the encoder is denoted as \emph{global memory} in Fig.~\ref{fig:frame}.

\noindent\textbf{Decoder}:
The decoder layer is also built upon the transformer architecture. Different from encoder layer, it contains an additional multi-head cross attention layer. The decoder transforms $N$ learnt positional embeddings (denoted as \emph{HOI queries} in Fig.~\ref{fig:frame}) into $N$ output embeddings. They are then decoded into HOI instances by the following MLP, which will be detailed in next section. In general, the decoder has three inputs, one is the global memory from encoder, one is HOI queries, and one is positional encoding. For multi-head cross attention layer, the \emph{Value} comes from global memory directly. The \emph{Key} is the sum of global memory and the input position encoding. The \emph{Query} is the sum of input position encoding and the input HOI queries. For self-attention layer, all of the Query, Key, Value come from the HOI queries or the output of previous decoder layer. The output of the decoder is denoted as \emph{output embeddings} in Fig.~\ref{fig:frame}. This architecture design follows~\cite{detr}.

\noindent\textbf{MLP for HOI Prediction}:
We define each HOI instance as a quintuple of (human class, interaction class, object class, human box, object box). The output embedding for each HOI query is decoded into one HOI instance by several multi-layer perception (MLP) branches. Specifically, there are three one-layer MLP branches to predict the human confidence, object confidence and interaction confidence respectively, and two three-layer MLP branches to predict human box and object box. All one-layer MLP branches for predicting confidence use a softmax function. For human confidence branch, the output size is 2, implies the confidences for foreground and background. For object confidence branch and interaction confidence branch, the output size is $C+1$, which implies the confidences for all $C$ kinds of objects or verbs defined in the dataset plus one for background. For both human and object box branches, the output size is 4, implies the normalized center coordinates $(x_{c}, y_{c})$, height and width of the box.

\subsection{HOI Instance Matching}

\begin{figure}[t]
\centering
\includegraphics[scale=0.5]{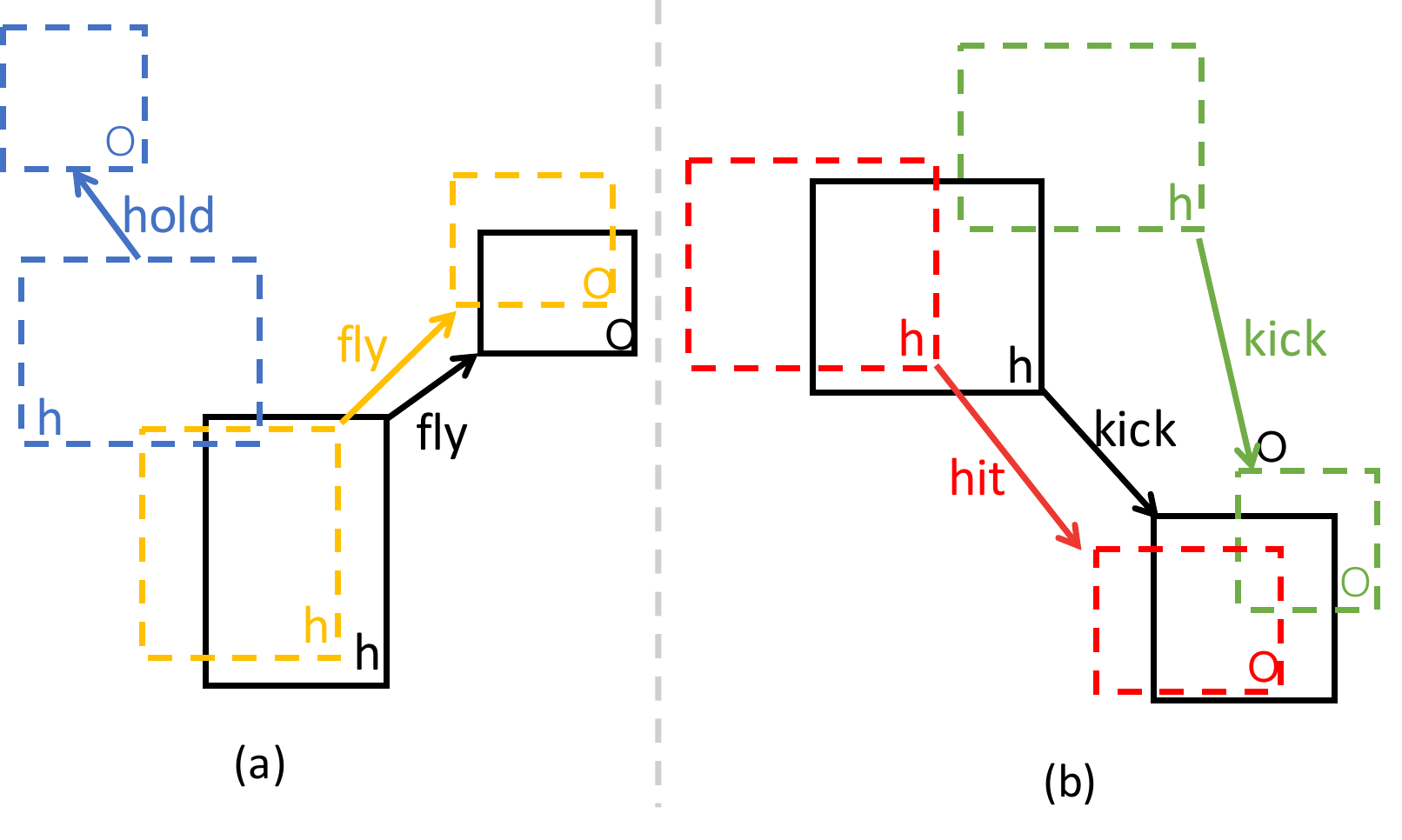}
\caption{Illustration of the matching strategy between HOI ground-truth (black) and prediction (other colors). An HOI instance is represented by a pair of boxes in the same color. $h$ and $o$ represent human and object respectively.}
\label{fig:me}
\end{figure}

% \noindent\textbf{Review of HOI detection.} 
The HOI instance is a quintuple of $\left(c_{h},  c_{r}, c_{o}, b_{h}, b_{o}\right)$, where $\left(c_{h}, c_{r}, c_{o} \right)$ denotes human, interaction  and object class confidence, $\left(b_{h}, b_{o}\right)$ is the bounding box of the human and object. Two-stage HOI detectors first predict the object proposals $ (c_{h}, b_{h}), (c_{o}, b_{o})$ with an object detector, then enumerate the detected (human, object) pairs to predict the $c_{r}$ by interaction classification. In other words, they are trying to approximate the following probability in a given dataset, 
\begin{equation}
\begin{aligned}
	p(h, r, o) &= p(h, o) p(r|h, o) \\
	&\approx p(h)  p(o)  p(r|h, o)
\end{aligned}
\end{equation}
where $p(h)$ and $p(o)$ indicate the confidence of human and object bounding box, respectively. $p(r|h, o)$ denotes the probability of interaction $r$ given human box $h$ and object box $o$, often implemented by a multi-stream interaction recognition model. In this method, the object detector and the interaction classifier are \emph{separately} optimized.

On the contrary, we treat HOI detection as a set prediction problem of bipartite matching between predictions and ground truth. Our method directly predicts the elements in HOI set and optimizes the proposed HOI matching loss in a unified way.

% Matching 
As shown in Fig.~\ref{fig:me}(a), suppose a ground truth (human, fly, object) is in the image, and the model  predicts two HOI instances: the yellow one (human, fly, object), and the blue one (human, hold, object). The yellow one not only predict the interaction correctly but localize the human and object more accurately as well. To minimize the matching cost, it is more suitable to assign the black one to the yellow one, and assign $\emptyset$ (implies nothing) to the blue one. 
% 差一句
A precise and complete matching strategy is formulated in the following.

%However, it is hard to divide the predictions distinctly  during the training process(e.g., the Fig.~\ref{fig:me}(b)). Therefore, a precise and complete matching process should be formulated.

Assume the model outputs a fixed-size set of $N$ predictions, where $N$ is chosen to be larger than the number of HOI relations in one image. Let us denote the set of predicted HOIs as $P={p^i, i=1,2,...,N}$, the set of ground truth HOIs as $G={g^i, i=1,2,...,M, \emptyset ,... ,\emptyset }$, where $M \leq N$. $M$ represents the number of ground truth in an image. By padding $\emptyset$ to the ground truth set, we make the length of two sets equal.

We denote the matching as an injective function: $\sigma_{G \rightarrow P}$, where $\sigma(i)$ is the index of predicted HOI assigned to the $i$-th groundtruth. The matching cost function is defined as:
\begin{equation}
	\mathcal{L}_{cost} = \sum_{i}^{N} \mathcal{L}_{\operatorname{match}}\left(g^{i}, p^{\sigma(i)}\right)
\end{equation}
where $\mathcal{L}_{\operatorname{match}}\left(g^{i}, p^{\sigma(i)}\right)$ is a matching cost between ground truth $g^{i}$ and prediction $p^{\sigma(i)}$.

In each step of training, we should first find an optimal one-to-one matching between the ground truth set and the current prediction set. We design the following matching cost for HOI: 

\begin{small}
%\begin{equation}
%\mathcal{L}_{\operatorname{match}}\left(g^{i}, p^{\sigma(i)}\right)=\sum_{j \in { h,o,r } }{\alpha_j \mathcal{L}_{\operatorname cls}^i} + \sum_{k \in {h,o }}{\beta_k \mathcal{L}_{\operatorname box}^i}
%\end{equation}
\begin{equation}
\mathcal{L}_{\operatorname{match}}\left(g^{i}, p^{\sigma(i)}\right)= \beta_{1}              \sum_{j \in { h,o,r } }{\alpha_j \mathcal{L}_{\operatorname{cls}}^j} + \beta_{2} \sum_{k \in {h,o }}{ \mathcal{L}_{\operatorname{box}}^{k}  }
\label{eq:matchcost}
\end{equation}
\end{small}
  where $\mathcal{L}_{\operatorname{cls}}^j = \mathcal{L}_{\operatorname{cls}}\left( g^{i}_{j}, p^{\sigma\left(i\right)}_{j}\right)$,  $j \in {h, o, r }$ represents human, object, and interactions, $g^{i}_{j}$ denotes the category label of $j$ on ground-truth   $g^{i}$. We use standard softmax cross entropy loss in the paper.  $\mathcal{L}_{\operatorname{box}}^{k}$ is box regression loss for human box and object box, the weighted sum of GIoU~\cite{giouloss} loss and $L_1$ loss are used.
$\alpha$ and $\beta$ are hyper-parameters of loss weights, which will be discussed later in ablation study. 

We use Hungarian algorithm~\cite{kuhn1955hungarian,detr} to solve the following problem to find a bipartite matching. 
\begin{equation}
\hat{\sigma}=\underset{\sigma \in \mathfrak{S}_{N} }{\arg \min } \mathcal{L}_{\operatorname{cost}}
\end{equation}
where $\mathfrak{S}_{N}$ denotes the one-to-one matching solution space.

After the optimal one-to-one matching between the groundtruth and predictions is found, the network loss is calculated between the matched pairs, using the same loss function as Eq.~\ref{eq:matchcost}. Although these two processes share the same formulation, the hyper-parameters of them are different theoretically and may have different optimal values. 
However, in practice, due to the considerable computation cost brought by large hyper-parameter search space, we made them  the same, just as DETR~\cite{detr} does.

Different from conventional HOI detection methods which optimize object detector and interaction classifier separately, the proposed HOI matching loss takes both the classification and localization into account. Human and object boxes will be produced simultaneously with their interactions.

%-----------------------------------
\section{Experiments }

\begin{table*}[]
\setlength{\tabcolsep}{3pt}
\begin{tabular}{lcccc|ccc|ccc}
\hline \hline
                                                               & \multicolumn{1}{l}{} & \multicolumn{1}{l}{} & \multicolumn{1}{l}{} & \multicolumn{1}{l}{} & \multicolumn{3}{|c}{Default}                                                                         & \multicolumn{3}{|c}{Known Object}                     \\
Methods                                                        & Feature Backbone     & Detector             & Pose             & Language         & Full$\uparrow$                  & Rare$\uparrow$                  & NonRare$\uparrow$               & Full$\uparrow$ & Rare$\uparrow$ & NonRare$\uparrow$ \\ \hline
 \textit{Two-stage methods} & & & & & & & & & & \\
Shen et al.~\cite{shen2018scaling}       & VGG-19               & COCO                 &                      &                      & 6.46                            & 4.24                            & 7.12                            &     -           &      -          &        -           \\
HO-RCNN~\cite{datahicodet}               & CaffeNet             & COCO                 &                      &                      & 7.81                            & 5.37                            & 8.54                            &  10.41              &      8.94          &         10.85          \\
InteractNet~\cite{kaiming2018detecting}  & ResNet-50-FPN        & COCO                 &                      &                      & 9.94                            & 7.16                            & 10.77                           &          -      &        -        &      -             \\
GPNN~\cite{qi2018learning}               & ResNet-101           & COCO                 &                      &                      & 13.11                           & 9.34                            & 14.23                           &       -         &      -          &       -            \\
iCAN~\cite{gao2018ican}                  & ResNet-50            & COCO                 &                      &                      & 14.84                           & 10.45                           & 16.15                           &     16.26           &     11.33           &    17.73               \\
PMFNet-Base~\cite{wan2019pose}           & ResNet-50-FPN        & COCO                 &                      &                      & 14.92                           & 11.42                           & 15.96                           &        18.83        &     15.30           &   19.89                \\
PMFNet~\cite{wan2019pose}                & ResNet-50-FPN        & COCO                 & $\checkmark$         &                      & 17.46                           & 15.65                           & 18.00                           &      20.34          &     17.47           &     21.20              \\
No-Frills~\cite{nofrills}                & ResNet-152           & COCO                 &                      & $\checkmark$         & 17.18                           & 12.17                           & 18.68                           &    -            &         -       &    -               \\
TIN~\cite{li2019tin}                     & ResNet-50            & COCO                 & $\checkmark$         &                      & 17.22                           & 13.51                           & 18.32                           &   19.38             &      15.38          &    20.57               \\
CHG~\cite{eccv2020hetegraph}             & ResNet-50            & COCO                 &                      &                      & 17.57                           & 16.85                           & 17.78                           &  21.00              &    20.74            &    21.08               \\
Peyre et al.~\cite{peyre2019detecting}   & ResNet-50-FPN        & COCO                 &                      & $\checkmark$         & 19.40                           & 14.63                           & 20.87                           &     -           &      -          &     -              \\
VSGNet~\cite{ulutan2020vsgnet}           & ResNet152            & COCO                 &                      &                      & 19.80                           & 16.05                           & 20.91                           &     -           &     -           &      -             \\
FCMNet~\cite{eccv2020keycues}            & ResNet-50            & COCO                 & $\checkmark$         & $\checkmark$         & 20.41                           & 17.34                           & 21.56                           &   22.04             &       18.97         &    23.12               \\
ACP~\cite{eccv2020actioncoprior}         & ResNet-152           & COCO                 & $\checkmark$         & $\checkmark$         & 20.59                           & 15.92                           & 21.98                           &        -        &    -            &    -               \\
Bansal et al.~\cite{bansal2020detecting} & ResNet-50-FPN        & HICO-DET             &                      & $\checkmark$         & 21.96                           & 16.43                           & 23.62                           &      -          &           -     &     -              \\
PD-Net~\cite{eccv2020polysemy}           & ResNet-152           & COCO                 &                      & $\checkmark$         & 20.81                           & 15.90                         & 22.28                          &      24.78          &   18.88             &       26.54            \\
PastaNet~\cite{li2020pastanet}           & ResNet-50            & COCO                 & $\checkmark$         & $\checkmark$         & 22.65                           & \textbf{21.17} & 23.09                           &       24.53         &       23.00         &     24.99              \\
VCL~\cite{eccv2020visualcomlear}         & ResNet101            & HICO-DET             &                      &                      & 23.63                           & 17.21                           & 25.55                           &     25.98           &    19.12            &  28.03                 \\
DRG~\cite{eccv2020dualgraph}             & ResNet-50-FPN        & HICO-DET             &                      & $\checkmark$         & 24.53                           & 19.47                           & 26.04                           &     27.98           &   \textbf{23.11}            &   29.43                \\  \hline \hline
\textit{One-stage methods} & & & & & & & & & & \\
UnionDet~\cite{eccv2020uniondet}         & ResNet-50-FPN        & HICO-DET             &                      &                      & 17.58                           & 11.52                           & 19.33                           &  19.76              &  14.68              &     21.27              \\ 
IPNet~\cite{wang2020irnet}               & Hourglass            & COCO                 &                      &                      & 19.56                           & 12.79                           & 21.58                           &    22.05            &      15.77          &     23.92              \\
% PPDM${}^{\dagger}$ ~\cite{liao2020ppdm}  & ResNet-101           & HICO-DET             &                      &                      & 19.75                           & 12.03                           & 22.05                           &                &                &                   \\
PPDM~\cite{liao2020ppdm}                 & Hourglass            & HICO-DET             &                      &                      & 21.73                          & 13.78                           & 24.10                           &  24.58              &   16.65             &           26.84        \\
\textbf{Ours}                                 & ResNet-50            & -                    &                      &                      & 23.46                          & 16.91                            & 25.41                          &    26.15            &     19.24           &  28.22                 \\
\textbf{Ours}                                 & ResNet-101           & -                    &                      &                      & \textbf{26.61}                           & 19.15                            & \textbf{28.84}                           &   \textbf{29.13}             &    20.98            &        \textbf{31.57}           \\
% \textbf{Ours${}^{\ddagger}$}                  & ResNet-50            & -                    &                      &                      & 24.67                           & 10.75                           & 28.82                           &                &                &                   \\
% \textbf{{}Ours${}^{\ddagger}$}              & ResNet-101           & -                    &                      &                      & \textbf{27.59} & 20.76                           & \textbf{29.63} &                &                &  \\   

\hline \hline
\end{tabular}
\caption{Comparison with the state-of-the-art methods on HICO-DET test set. For the Detector, COCO means that the detector is trained on COCO, while HICO-DET means that the detector is first trained on COCO and then fine-tuned on HICO-DET. Pose means that human pose feature extracted by pre-trained skeleton model. Language means that languages prior, e.g. words2vec.  Our methods achieve best performance on Full categories without introducing additional human pose feature, external language prior, and object detector.}
\label{tab:hico}
\end{table*}

\subsection{Experimental Setting}

\noindent\textbf{Datasets}:
We conduct experiments on HICO-DET~\cite{datahicodet} and V-COCO~\cite{datavcoco} benchmark to evaluate the proposed methods. HICO-DET consists of 47,776 images with more than 150K human-object pairs (38,118 images in training set and 9,658 in test set). It has 600 HOI categories over 117 interactions and 80 objects. Further, 600 HOI categories has been split into 138 Rare and 462 Non-Rare based on the number of training instances.
V-COCO is a subset of MS-COCO~\cite{lin2014microsoft}, consists of 5,400 images in the trainval dataset and 4946 images in test set. Each human is annotated with binary labels for 29 different action categories (five of them do not involve associated objects).  

\noindent\textbf{Evaluation Metric}:
Following the standard evaluation, we use the commonly used role \textit{mean average precision} (mAP) to examine the model performance for both datasets. An HOI detection is considered as true positive if and only if it localizes the human and object accurately (i.e. the \textit{Interaction-over-Union} (IOU) ratio between the predicted box and ground-truth is greater than 0.5) and predict the interaction correctly.

\noindent\textbf{Implementation Details}

\textit{Data Augmentation}: First, we adjust the brightness and contrast with a probability of 0.5 as  image level augmentation. Specifically, for both brightness and contract, a parameter is randomly chosen from the range [0.8, 1.2], meaning only slight change is performed to original image. Next, we use scale augmentation, scaling the input image such that the shortest side is at least 480 and at most 800 pixels while the longest at most 1333~\cite{wu2019detectron2}. And also, we use random flip with a probability of 0.5. Finally, we apply random crop augmentations: an image is cropped with probability 0.5 to a random rectangular patch followed by another scale augmentation to ensure its shape, it is noteworthy that if any box in a given ground truth human-object pair is outside the cropped patch, its label will be removed.

\textit{Training Settings}: The input image to the model is first scaled to [0, 1] and then normalized by channel-wise mean and std. The experiments are conducted on two popular backbone, ResNet-50 and ResNet-101. The models are trained with AdamW~\cite{adamw} setting the transformer’s learning rate to 1e-4, the backbone’s to 1e-5, and weight decay to 1e-4. The number of encoder layer and decoder layer are both set to 6, the number of HOI query is set to 100, and we use a COCO pre-trained DETR~\cite{detr} model to initialize the weights of both backbone and transformer encoder-decoder. The batch size for ResNet-50 is set to 16 while 8 for ResNet-101. All the models   are trained for 250 epochs with once learning rate decay at epoch 200. 
%All models are trained on an 8-GPU machine. 
Training our network takes 7 hours on 8 NVIDIA 2080TI GPU on V-COCO and 70 hours on HICO-DET. At test times, our model runs at 24 fps on a single 2080TI GPU.

\subsection{Comparisons with State-of-the-Art methods}
We report the main quantitative resutls in terms of $AP$ on HICO-DET in Table~\ref{tab:hico} and $AP_{role}$ on V-COCO in Table~\ref{tab:vcoco}.

% 实验分析
For the HICO-DET dataset, our method compares against state-of-the-art algorithms. We achieve $4.88\%$ point gain over one-stage methods~\cite{liao2020ppdm} on Full categories, especially $5.37\%$ point on Rare categories. 
% 1. human pose 很重要， Language 很重要, 最后去提下rare的掉点
Compared with those two-stage methods, our method achieves best performance on Full categories without introducing additional human pose feature and languages prior, which shows great potential of our method.
Meanwhile, we humbly regard how to improve performance on Rare due to the long-tail distribution of HOI detection as a significant future work in our framework. 

% As can be seen from Table.~\ref{tab:hico}, additional language prior is significant for Rare categories. 
For the V-COCO dataset, our method also achieves the competitive performance compared with state-of-the-art methods without introducing external data, e.g. human pose, HICO-DET data, language prior knowledge. We obtain $1.9\%$ point gain over previous one-stage method~\cite{wang2020irnet}.

\begin{table}[t]
\setlength{\tabcolsep}{1.6pt}
\label{tab:vcoco}
\begin{tabular}{llccc}
\hline \hline
Methods & Backbone & Pose & Language & AProle \\ \hline
\textit{Two-stage methods} \\
VSRL~\cite{datavcoco} & ResNet-50-FPN & &  & 31.8 \\
InteractNet~\cite{kaiming2018detecting} & ResNet-50-FPN &  &  & 40.0 \\
GPNN~\cite{qi2018learning} & ResNet-101 & &  & 44.0 \\
RPNN~\cite{zhou2019relation} & ResNet50 & $\checkmark$ & & 47.5 \\
VCL~\cite{eccv2020visualcomlear} & ResNet101 &  & & 48.3 \\

TIN${}^*$~\cite{li2019tin} & ResNet-50 & $\checkmark$ & & 48.7 \\

Zhou et al.~\cite{zhou2020cascaded} & ResNet-50 & $\checkmark$ & & 48.9 \\

PastaNet~\cite{li2020pastanet} & ResNet-50 & $\checkmark$  & $\checkmark$ & 51.0 \\

DRG~\cite{eccv2020dualgraph} & ResNet-50-FPN &  & $\checkmark$ & 51.0 \\

VSGNet~\cite{ulutan2020vsgnet} & ResNet-152 &  &  & 51.8 \\
CHG~\cite{eccv2020hetegraph} & ResNet-50 & & & 52.7 \\
PMFNet~\cite{wan2019pose} & ResNet-50-FPN & $\checkmark$ &  & 52.0 \\ 

PD-Net~\cite{eccv2020polysemy} & ResNet-152 & $\checkmark$ &  & 52.6 \\

FCMNet~\cite{eccv2020keycues} & ResNet-50 & $\checkmark$ & $\checkmark$ & 53.1 \\
ACP${}^*$~\cite{eccv2020actioncoprior} & ResNet-152 & $\checkmark$ & $\checkmark$ & \textbf{53.2} \\

\hline \hline
\textit{One-stage methods} \\
UnionDet~\cite{eccv2020uniondet} & ResNet-50-FPN &  &  & 47.5 \\
IPNet~\cite{wang2020irnet} & Hourglass-104 & & & 51.0 \\
IPNet${}^*$~\cite{wang2020irnet} & Hourglass-104 & &  & 52.3 \\

% \textbf{Ours} & ResNet-50  &  &  & 52.8 \\ 

\textbf{Ours} & ResNet-101  &  &  & \textbf{52.9} \\ 

\hline \hline
\end{tabular}
\caption{Comparisons of the state-of-the-art on V-COCO test set. Pose denotes whether human skeleton feature has been introduced and Language denotes external-language prior. Character ${}^*$ indicates that HICO-DET training data was incorporated into training data.}
\label{tab:vcoco}
\end{table}

\begin{table}[t] 
\centering
	\begin{subtable}[h]{0.35\textwidth}
		\centering
		\begin{tabular}{ccccc}
\hline
$\alpha_{r}$ &  $\alpha_{o}$ & Full$\uparrow$ & Rare$\uparrow$ & NonRare$\uparrow$ \\ \hline
 1.0 & 1.0 & 16.97 & 11.65  & 18.55 \\
 1.0 & 2.0 & 17.49 & 11.47  & 19.29 \\
 2.0 & 1.0 & 17.85 & 13.54  & 19.14 \\ \hline
\end{tabular}	
		\caption{Varying $\alpha$ for classification loss}
		\label{tab:ablationlossweight}
	\end{subtable}
	\hfill
	\begin{subtable}[h]{0.35\textwidth}
		\centering
		\begin{tabular}{ccccc}
\hline
 $\beta_{1}$ & $\beta_{2}$ & Full$\uparrow$ & Rare$\uparrow$ & NonRare$\uparrow$ \\ \hline
 1.0 & 1.0 & 16.97 & 11.65  & 18.55 \\
 1.0 & 2.0 & 16.20 & 9.54  & 18.19 \\
 2.0 & 1.0 & 18.46 & 13.41  & 19.97 \\ \hline
\end{tabular}
		\caption{Varying $\beta$ for match cost}
		\label{tab:ablationmatchcost}
	\end{subtable}
	\hfill
	\begin{subtable}[h]{0.35\textwidth}
		\centering
		\begin{tabular}{ccccc}
\hline
scale & crop & Full$\uparrow$ & Rare$\uparrow$ & NonRare$\uparrow$ \\ \hline
 &  &16.97 & 11.65  & 18.55 \\
 & $\checkmark$ & 22.05 & 15.20 & 24.09 \\
$\checkmark$ &  & 21.26 & 14.61 & 23.25 \\
$\checkmark$ & $\checkmark$ & 22.36 & 15.47 & 24.42 \\ \hline
\end{tabular}
		\caption{Data augmentation during training process}
		\label{tab:ablationdataaug}
	\end{subtable}
	\caption{Ablation experiments for HOI Transformer. All models use ResNet-50 backbone to extract feature;  number of queries is set to 100; batch size is set for 16; trained for 250 epochs; learning rate decay from 1e-4 to 1e-5 at epoch 200.}
	\label{tab:ablationall}
\end{table}

\subsection{Ablation Study}

% 1. 总领的话

In our ablation study, we explore how the matching strategy, loss weight, and data augmentation influence the final performance. 
The ablation experiments are conducted with ResNet-50 backbone models, and the models are trained for 250 epochs with once learning rate decay at epoch 200. 
The number of encoder layer and decoder layer are both set to 6, the number of HOI query is set to 100, and the batch size is set to 16. We use a COCO pre-trained DETR model to initialize the weights of both backbone and transformer encoder-decoder.

Our baseline set all the loss weight hyper-parameters to 1.0, and uses only brightness, contrast and random flip augmentation.

\noindent\textbf{Matching Strategy}:
Consider the situation in Fig.~\ref{fig:me}(b), black boxes indicate a ground truth pair, green boxes indicate a predicted pair, and red boxes indicate another predicted pair.
For simplicity, we assume that both human and object boxes are in the right place and the object class category is correct. Then we can see, for the red pair, both human and object boxes have higher overlap than the black ones, but its interaction prediction `hit' is wrong, note that the interaction label of ground truth is `kick'. For the green pair, the interaction prediction `kick' is right, while its human box is obviously far from the ground truth. So in this case, which pair to match with the ground truth is confused, location first or category first? We conduct ablation study to find the relative importance in matching. In Eq.~\ref{eq:matchcost}, $\beta_{1}$, $\beta_{2}$ dominates the weight of classification and localization respectively. As shown in Table.~\ref{tab:ablationmatchcost}, the best result is obtained under $\beta_{1}=2.0, \beta_{2}=1.0$, which reflects that classification plays a more important role than localization during the matching process.

\noindent\textbf{Loss Ablation}: The human/object/interaction are $2/81/117$-category classification problems in HICO-DET. `human' has sufficient training data because it appears in each HOI instance. Therefore, we assume human classification as the simplest one and set $\alpha_h = 1$. We conduct experiments to evaluate the relative importance of `object' and `interaction' in our experiments. In Eq.~\ref{eq:matchcost}, $\alpha_{o}$ and $\alpha_{r}$  dominate the weight of object and interaction respectively in training loss. As shown in Table.~\ref{tab:ablationlossweight}, our method obtains best result when $\alpha_{r}=2.0$ and $\alpha_{o}=1.0$, indicating that interaction tends to be more important than object in our framework.

% 加一些insight
\noindent\textbf{Data Augmentation}: 
%We evaluate two ways of data augmentations for our methods, multi-scale training and random crop.
We mainly study two kinds of data augmentation in our experiments: multi-scale training and random crop. We conduct ablation experiments on combination of them, results can be found in Table.~\ref{tab:ablationdataaug}.
%To improve the scale-invariance and shift-invariance of transformer encoder-decoder, we introduced two ways of data augmentations for our methods, multi-scale training and random crop.  
%Table~\ref{tab:ablationdataaug} shows the comparisons with baselines and multi-scale, random crop in training process.
Considerable improvements have been made, multi-scale training attains $4.29\%$ point gain on Full categories and random crop achieves $5.08\%$, and the combination of them gets even better results, mainly because these two augmentations help the attention layers to learn scale-invariant and shift-invariant features much easier on a small dataset. 

%which may be because the data augmentation facilitate the learning of scale-invariance and shift-invariance   self-attention layer in our HOI transformer.
 
\begin{figure}[h]
\centering
\includegraphics[scale=0.2]{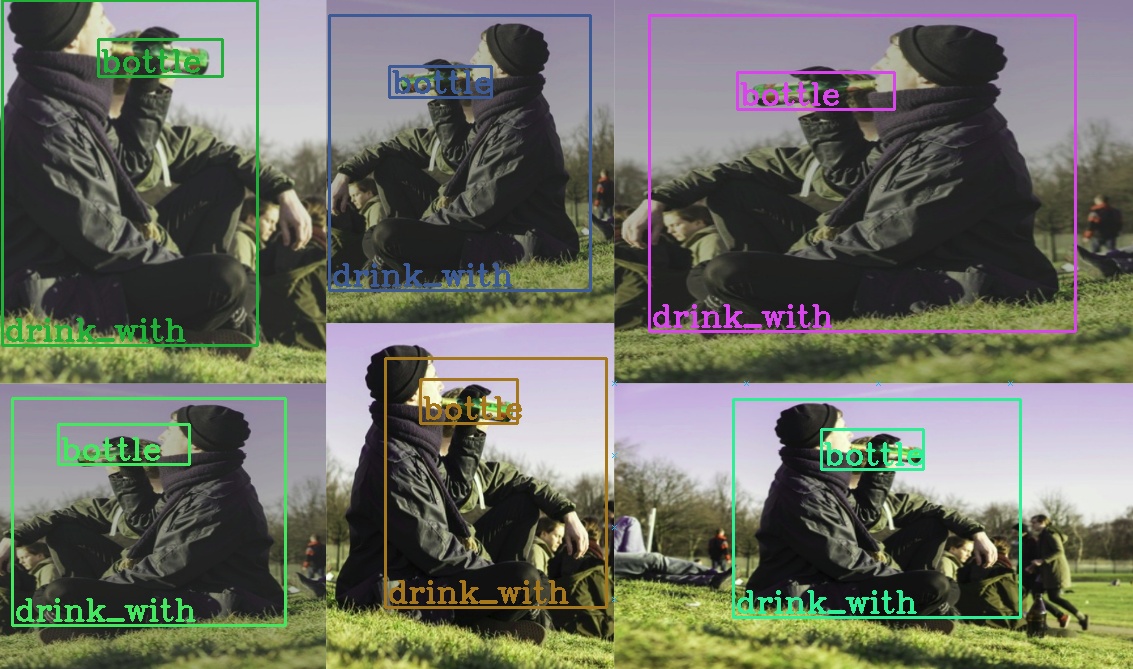}
\caption{Out of distribution generalization for HOI prediction. Even though no images  in the training set has  more than 3 `drink with bottle' HOIs, our method generalize well  on the synthetic image with 6  of them.}
\label{fig:6drink}
\end{figure}

\begin{figure*}[ht]
\centering
\includegraphics[width=\textwidth,height=0.35\textheight]{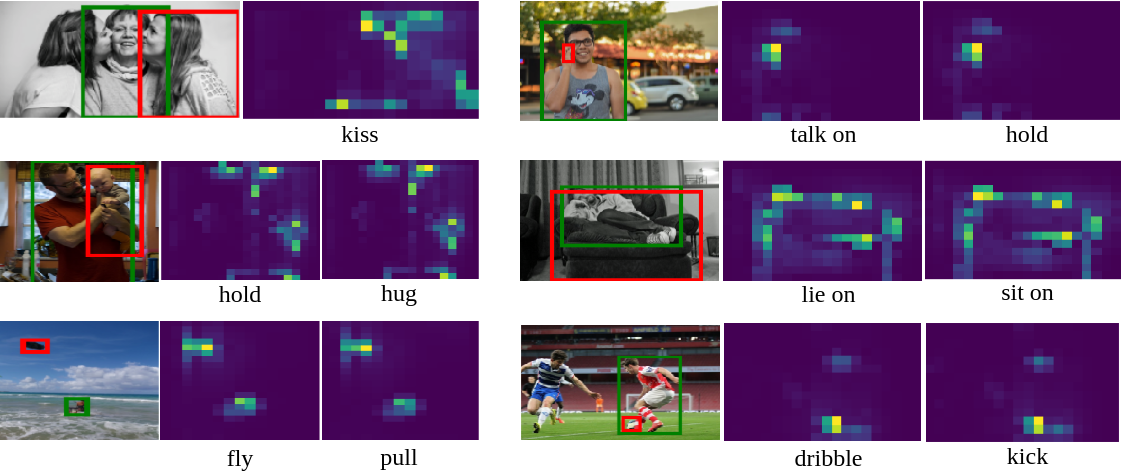}
\caption{Visualization of attention map in decoder for predicted HOI instance (images from HICO-DET dataset). As can be seen from the figure, our method attends to the discriminative part (e.g. telephone for \emph{talk on}, human face for \emph{kiss}) and  can capture long-distance interaction (e.g. \emph{fly} and \emph{pull}).   
Moreover, it can be seen from the figure that different interaction categories share little common pattern, which suggested a unified definition for all interaction proposal may be sub-optimal.}
\label{fig:attenion}
\end{figure*}

\subsection{Discussion}
% --------- 引入 ------- 提我们的data point是image
We formulate HOI detection as a set prediction problem and directly predict instances in an end-to-end manner. 
So, compared with previous methods, which are  trained on HOI instance level, treating human-object pair as data point, our method regards a whole image as data point. There seems some differences. Here we briefly discuss our understanding on two aspects.

\emph{Image-level data point.}
First, the training loss for interaction in our method is the sum of classification loss between each matched (pred, GT) pair. It is relevant to the number of positive samples, which is consistent with prior instance-level work. 
Second, random crop augmentation provides sufficient training instances. For example, let's think random crop in an extreme way, each time only one positive sample is cropped, then it is close to prior work.

\emph{Out of distribution test.}  
% We evaluate our method on a synthetic image containing $6$ (human, drink with,  bottle) HOI instances(at most $3$ in training data). The ideal result indicates that the model can learn to recognize HOI  with good generalization.
Noting that in training data there are at most 3 `drink with bottle' HOIs in a single image, we create a synthetic image containing 6 `drink with bottle' HOIs, which is out of distribution. As shown in Fig.~\ref{fig:6drink}, the result indicates that the model can learn to recognize HOIs with good generalization.

\subsection{Qualitative Analysis}
 As can be seen from Fig.~\ref{fig:attenion}, we visualize the decoder attention map for predicted HOI instances. The interaction heatmap highlights both the human and object area, meaning that our model reasons about the relations between human and object from a more global image context, not focusing on human or object only. It is obvious that decoder has ability to find the discriminative part for the interaction category. The model can predict different instances based on similar attention heatmap, which implies that, the MLP of the model have the ability to tell from fine-grained interaction features. Meanwhile,  some local area with relatively higher attention may indicate the localization (boundaries) of human or object, because the visualized attention map is immediately followed by the MLP head for classification as well as regression.
 Moreover, it can be seen from the figure that different interaction categories share little common pattern, which suggests that the empirically unified definition  of interaction proposal in one-stage methods, i.e. interaction point/box, is sub-optimal. And thanks to the large receptive field of attention layers, our model can easily handle long distance interaction, e.g. fly kite.

% As can be seen from Fig.~\ref{fig:attenion}, we visualize the decoder attention map for predicted HOI instance. It is obvious that decoder has ability to find the discriminative part for the interaction category. Moreover, it can be seen from the figure that different interaction categories share little common pattern, which suggests that the unified definition of interaction proposal in one-stage methods is sub-optimal. And thanks to the large receptive field of attention layers, we can easily handle long distance interaction, e.g. fly kite.
% 
% From each image-heatmap pair we can see, the model can
% 
% \noindent\textbf{Discussion.} As shown in Table~\ref{tab:hico} and Table~\ref{tab:vcoco}, HOI Transformer outperforms other state-of-the-art methods without introducing additional feature, which shows great potential of our method. Meanwhile, we humbly admit that HOI Transformer need to improve performance on rare-set due to the long-tail distribution of HOI detection.

% Performance on HICO-DET rare-set is  slightly lower than other one-stage methods. It is suggested that human pose feature and language prior is significant for rare-set. It still leaves great improvements for our method. 

% performing of our HOI Transformer obtains considerable performance gain on NonRare categories($6.2\%$ point), while lower on Rare subset.  It is obvious that recent methods obtain $$ additional human pose and language prior     

\section{Conclusion}
In this paper, we propose a novel HOI Transformer to directly predict the HOI instances in an \emph{end-to-end} manner. Our core idea is to build a transformer encoder-decoder architecture to directly predict HOI instances, and a quintuple matching loss for HOI to enable supervision in a unified way.
We validate the proposed method on two challenging HOI benchmark and achieve a considerable performance boost over state-of-the-art results. It is worth noting that our method not only abandon the additional features, but desert the complex post processing as well. Moreover, based on the attention map of the decoder, we found that our model has ability to dynamically attains the discerning feature for different HOI queries. We hope our method will be useful for human activity understanding research.

% ~\\
% \noindent\textbf{Acknowledgements.} This research was supported by China's ``scientific and technological innovation 2030 - major projects" (No. 2020AAA0104400)

\section*{Acknowledgement}
This research is supported by China's `scientific and technological innovation 2030 - major projects' (No. 2020AAA0104400).

\clearpage
% \section*{Appendix}
\section*{Appendix A: Illustration of Inference Process}
With parallel decoding~\cite{oord2018parallel,gu2017non,ghazvininejad2019mask}, the network can simultaneously output a group of $N$ HOI instances. In training phase, the $N$ output HOI instances first find an optimal one-to-one matching to the $N$ ground truth, and then get the loss between matched pairs to optimize the network, as shown in the right part of  Fig.~\ref{fig:frame}. While in inference phase, no one-to-one matching is needed, the network directly outputs $N$ HOI instances. 

Regardless of the parallel decoding technology used in our architecture, the inference process can be comprehended in an auto-regressive way~\cite{ng}. A high level global memory full of rich context is first extracted by the encoder. At each time step, the decoder outputs a new HOI instance considering both the global feature and the previous detected HOI instances. Only the undetected HOI instances will be figured out at each time step. The most likely HOI candidates are encouraged to be produced at first. 

\begin{figure}[h]
\centering
\includegraphics[scale=0.5]{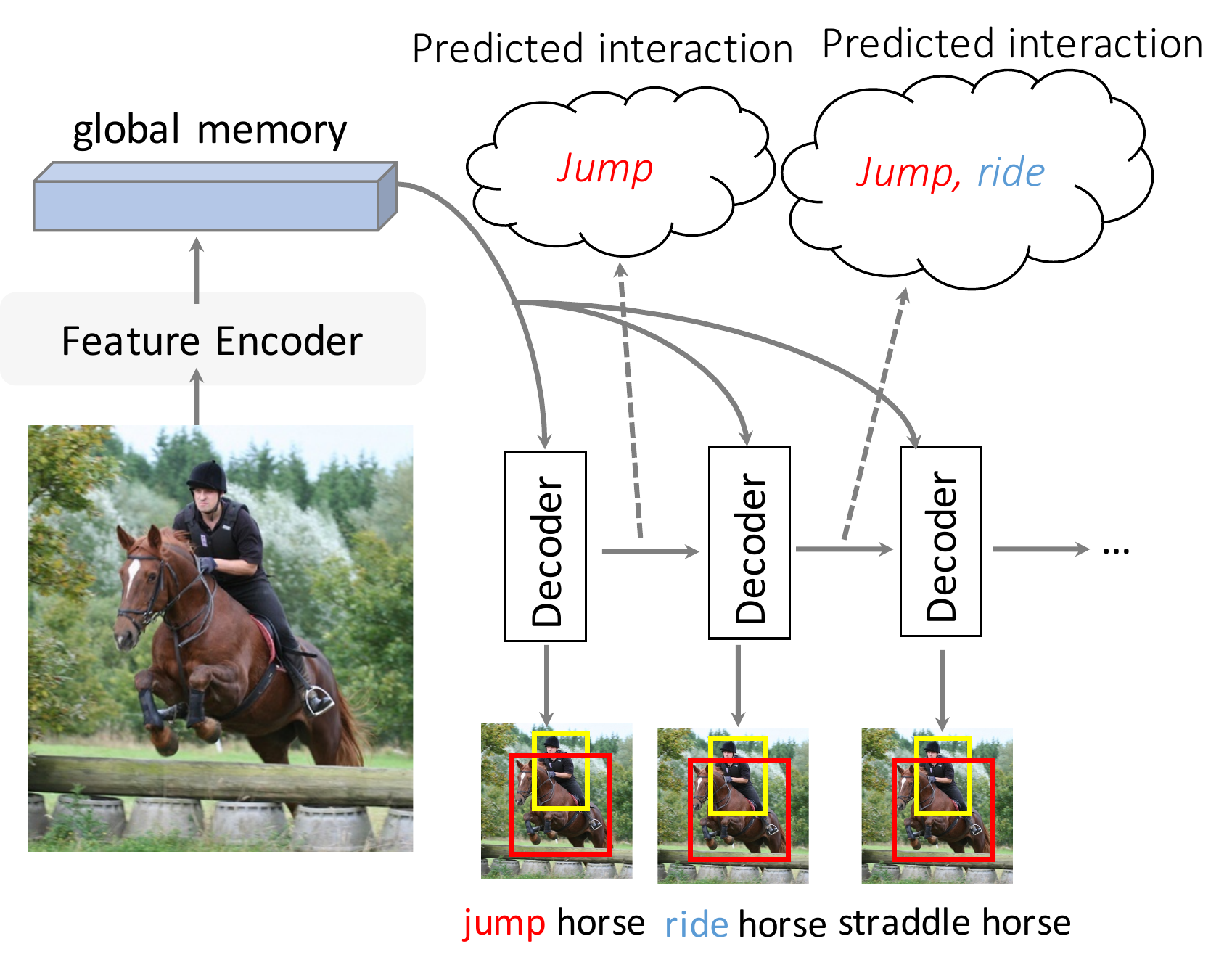}
\caption{Examples of decoder inference process. At each time step, the decoder outputs a new HOI instance considering both the global memory  and the previously detected HOI instances.}
\label{fig:menmian}
\end{figure}

As shown in Fig.~\ref{fig:menmian}, global memory from an image containing person ride/jump/straddle horse is first extracted. At time step 1, the most likely interaction `jump horse' is produced. When it comes to time step 2, the model looks through the global memory and the previous output and find that `jump horse' has already been predicted. So, it outputs the second most likely interaction `ride horse'. It keeps producing HOI instances until there is a stop token, i.e. the confidence is lower than a predefined threshold, or has reached the maximum number.

{\small
\bibliographystyle{ieee_fullname}
\bibliography{egbib}
}

\end{document}